\documentclass[11pt]{article}

\usepackage[utf8]{inputenc}
\usepackage[T1]{fontenc}
\usepackage{amsmath, amssymb, amsthm}
\usepackage{graphicx}
\usepackage{hyperref}
\usepackage{cite}
\usepackage{booktabs}
\usepackage{multirow}
\usepackage{url}

\usepackage[margin=1in]{geometry}

\title{Deep Learning-Based Early-Stage IR-Drop Estimation via CNN Surrogate Modeling}

\author{
    Ritesh Bhadana\\Gurugram University, India. \\{riteshbhadanaa@gmail.com}
}

\date{}

\begin{document}

\maketitle

\begin{abstract}
IR-drop is a critical power integrity challenge in modern VLSI designs that can cause timing degradation, reliability issues, and functional failures if not detected early in the design flow. Conventional IR-drop analysis relies on physics-based signoff tools, which provide high accuracy but incur significant computational cost and require near-final layout information, making them unsuitable for rapid early-stage design exploration.

In this work, we propose a deep learning-based surrogate modeling approach for early-stage IR-drop estimation using a CNN. The task is formulated as a dense pixel-wise regression problem, where spatial physical layout features are mapped directly to IR-drop heatmaps. A U-Net-based encoder-decoder architecture with skip connections is employed to effectively capture both local and global spatial dependencies within the layout.

The model is trained on a physics-inspired synthetic dataset generated by us, which incorporates key physical factors including power grid structure, cell density distribution, and switching activity. Model performance is evaluated using standard regression metrics such as Mean Squared Error (MSE) and Peak Signal-to-Noise Ratio (PSNR). Experimental results demonstrate that the proposed approach can accurately predict IR-drop distributions with millisecond-level inference time, enabling fast pre-signoff screening and iterative design optimization.

The proposed framework is intended as a complementary early-stage analysis tool, providing designers with rapid IR-drop insight prior to expensive signoff analysis. The implementation, dataset generation scripts, and the interactive inference application are publicly available at: \url{https://github.com/riteshbhadana/IR-Drop-Predictor}. The live application can be accessed at: \url{https://ir-drop-predictor.streamlit.app/}.
\end{abstract}


\section{Introduction}
\label{sec:introduction}

IR-drop is a critical power integrity issue in modern VLSI designs, caused by resistive losses in the power delivery network. Excessive IR-drop can lead to timing degradation and functional failures, making its early detection essential for reliable chip design. As technology scales and circuit density increases, accurately analyzing IR-drop has become more challenging.

Conventional IR-drop analysis relies on physics-based simulation methods, which are highly accurate but computationally expensive. These methods are typically applied at late design stages and require significant runtime, making them unsuitable for fast design iteration during early physical planning.

To enable rapid early-stage IR-drop estimation, this work proposes a deep learning-based surrogate model that predicts IR-drop distributions directly from layout-related features. The problem is formulated as a pixel-wise regression task, where spatial feature maps such as power grid structure, cell density, and switching activity are mapped to an IR-drop heatmap.

A U-Net-based CNN is used due to its ability to capture both local and global spatial patterns through an encoder-decoder structure with skip connections. The model is trained using a physics-inspired synthetic dataset, allowing it to learn realistic IR-drop behavior without requiring costly signoff simulations during inference.

The proposed approach provides fast and efficient IR-drop prediction, enabling early identification of high-risk regions and supporting rapid design exploration. While not a replacement for detailed physics-based tools, it serves as a practical pre-signoff solution to guide early power integrity analysis.

\section{Problem Background}
\label{sec:background}

IR-drop is the reduction in supply voltage caused by current flowing through resistive elements of the power delivery network. With increased circuit density, switching activity, and routing constraints in advanced technology nodes, IR-drop has become a major challenge in modern VLSI design. Excessive IR-drop can result in voltage droop, timing violations, and functional failures.

Accurate IR-drop analysis requires complex physics-based modeling of current flow and voltage distribution, which is computationally expensive and typically applied at late design stages. This limitation creates a need for fast and scalable early-stage estimation methods. Data-driven surrogate models offer a practical alternative by approximating IR-drop behavior without explicitly solving physical equations.

\section{Related Work and Comparison with Prior Approaches}
\label{sec:related}

IR-drop analysis has traditionally been addressed using physics-based simulation methods that model current flow and voltage distribution across the power delivery network. These approaches rely on solving large-scale electrical equations using finite-element or SPICE-based solvers and are widely adopted in industrial design flows. While highly accurate, such methods are computationally expensive and unsuitable for fast iteration during early design stages.

In recent years, machine learning techniques have been explored to accelerate power integrity analysis. Some prior works apply regression models or shallow neural networks to estimate IR-drop using handcrafted features or reduced representations. Other studies investigate deep learning approaches for layout-aware prediction, often focusing on specific design stages or limited spatial resolution.

More recent research has introduced CNN for spatial IR-drop prediction, treating the problem as an image-to-image regression task. These methods demonstrate that CNNs can learn spatial voltage patterns efficiently. However, many existing approaches either depend on signoff-level data, require complex feature engineering, or lack a clear formulation suitable for early-stage design exploration.

\subsection{Difference from Prior Work}

The key differences between this work and existing approaches are summarized as follows:

\begin{itemize}
    \item \textbf{Problem Formulation:} This work formulates IR-drop estimation explicitly as a pixel-wise regression problem, enabling dense spatial prediction rather than coarse scalar estimation
    \item \textbf{Model Architecture:} A U-Net-based CNN is employed to preserve spatial resolution using skip connections, which improves hotspot localization compared to standard CNN architectures
    \item \textbf{Data Strategy:} The model is trained on a physics-inspired synthetic dataset, allowing learning of realistic IR-drop behavior without requiring expensive signoff simulations
    \item \textbf{Design Stage Focus:} Unlike many prior works that target late-stage analysis, this approach is designed for early-stage power integrity screening, where fast feedback is critical
    \item \textbf{Practical Deployment:} The proposed model is integrated into an interactive inference pipeline, demonstrating usability beyond offline analysis
\end{itemize}

Overall, this work extends prior research by combining a spatially expressive deep learning architecture with a physics-informed data formulation, providing a fast and scalable surrogate model for early IR-drop estimation.

\section{IR-Drop Theory}
\label{sec:theory}

IR-drop is fundamentally governed by electrical resistance and current flow in the power delivery network. At a basic level, the voltage drop across a conductor is described by Ohm's law:
\begin{equation}
V_{\text{drop}} = I \times R
\label{eq:ohm}
\end{equation}
where $I$ represents the current drawn by the circuit and $R$ denotes the effective resistance of the power grid path. As current demand increases or power grid resistance becomes higher, the resulting voltage drop also increases.

In large-scale integrated circuits, IR-drop is not confined to a single conductor but is distributed across a complex, multi-layer power grid. The voltage distribution over such a grid is governed by continuous electrical models that describe current flow and conductivity. At a conceptual level, this behavior can be expressed using a partial differential equation of the form:
\begin{equation}
\nabla \cdot (\sigma \nabla V) = -J
\label{eq:pde}
\end{equation}
where $V$ is the voltage, $\sigma$ represents electrical conductivity, and $J$ denotes current density. This equation models how voltage varies spatially due to resistive losses and current consumption across the grid.

Solving this equation accurately requires numerical methods over large spatial domains, which leads to high computational cost. As a result, full physics-based IR-drop analysis is typically performed only at later stages of the design flow. In this work, rather than explicitly solving these equations during inference, the deep learning model learns to approximate their effects by mapping layout-level features directly to IR-drop distributions.

\section{Physical Relevance of Model Inputs}
\label{sec:inputs}

The input features used in this work are chosen to reflect the key physical factors that influence IR-drop in integrated circuits. IR-drop is fundamentally driven by the interaction between current demand and the resistance of the power delivery network, both of which are spatially varying across the chip.

\subsection{Power Grid Map}

The power grid map shows the strength of the power delivery wires. If the metal wires are too thin or spread out, they have high resistance, which makes it harder for power to reach its destination, leading to increased voltage drop under load.

\subsection{Cell Density Map}

The cell density map shows where the circuit components are packed together. Areas with too many components draw a lot of current, which puts a heavy load on the power grid and increases the likelihood of IR-drop due to higher static and dynamic power consumption.

\subsection{Switching Activity Map}

The switching activity map tracks how often circuit components turn on and off. High activity creates sudden spikes in power use, which can cause the voltage to dip suddenly in those spots, leading to localized voltage droop.

By analyzing these three maps together, we get a complete picture of how power flows through the chip. Because these maps are spatial—meaning they look like images—the problem is well-suited for a CNN. Instead of performing slow, complex numerical computations, the CNN learns the complex patterns in these maps to quickly predict exactly where voltage will drop.

\begin{figure}[t]
\centering
\begin{minipage}{0.3\textwidth}
    \centering
    \includegraphics[width=\textwidth]{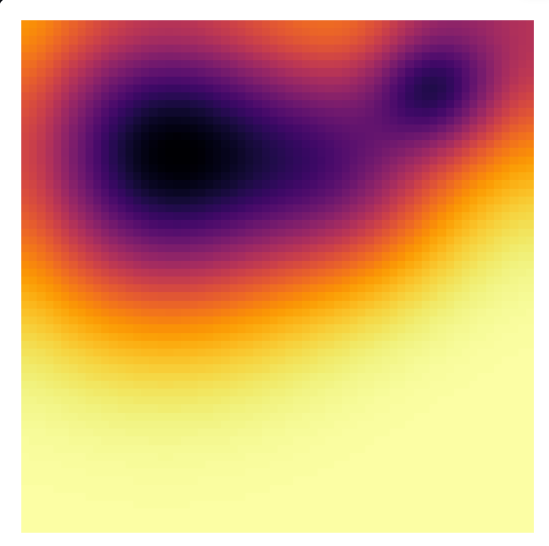}
    \small Heatmap of Power Grid
\end{minipage}
\hfill
\begin{minipage}{0.3\textwidth}
    \centering
    \includegraphics[width=\textwidth]{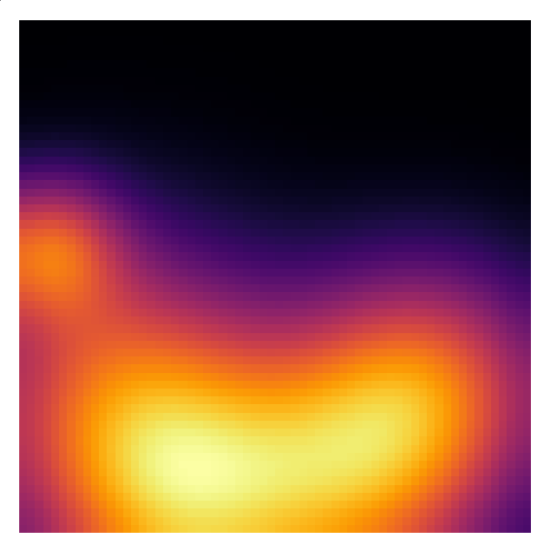}
    \small Heatmap of Cell Density
\end{minipage}
\hfill
\begin{minipage}{0.3\textwidth}
    \centering
    \includegraphics[width=\textwidth]{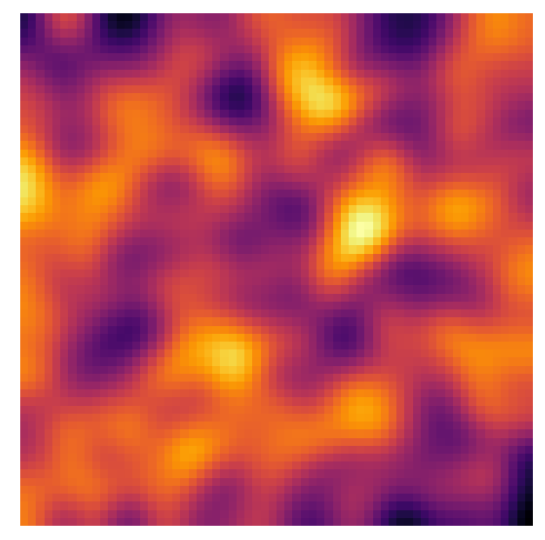}
    \small Heatmap of Switching Activity
\end{minipage}
\caption{Three spatial feature maps capturing the key physical factors influencing IR-drop distribution: power grid strength (left), cell density (middle), and switching activity (right).}
\label{fig:input_features}
\end{figure}

\section{Dataset Description}
\label{sec:dataset}

The dataset used in this work consists of spatial feature maps stored in NumPy (.npy) format, enabling efficient loading and processing of large numerical arrays. Each sample represents a localized region of a chip layout and is encoded as two-dimensional matrices of size $64 \times 64$.

For every sample, four maps are provided:

\begin{itemize}
    \item \textbf{Power Grid Map:} Represents the effective strength of the power delivery network, indirectly modeling local resistance variations
    \item \textbf{Cell Density Map:} Encodes the spatial distribution of logic cells, serving as a proxy for static and dynamic current demand
    \item \textbf{Switching Activity Map:} Captures signal transition activity, reflecting dynamic current fluctuations
    \item \textbf{IR-Drop Map (Ground Truth):} Represents the corresponding voltage drop distribution across the same region
\end{itemize}

All values are normalized to the range $[0,1]$ to ensure numerical stability during training. The dataset contains diverse spatial patterns, including both smooth global voltage gradients and localized hotspot regions, allowing the model to learn a wide range of IR-drop behaviors.

The use of grid-based numerical representations enables the problem to be formulated as a pixel-wise regression task, making it well suited for cnn architectures.

\begin{table}[t]
\centering
\caption{Input-output mapping for IR-drop prediction}
\label{tab:dataset}
\small
\begin{tabular}{p{1.5cm}p{3.5cm}p{1.5cm}p{3cm}p{3.5cm}}
\toprule
\textbf{Category} & \textbf{File Name} & \textbf{Shape} & \textbf{Physical Meaning} & \textbf{Role in Model} \\
\midrule
Input & input\_power\_grid.npy & $64 \times 64$ & Local power grid strength and effective resistance & Encodes resistance-related voltage drop behavior \\
\midrule
Input & input\_cell\_density.npy & $64 \times 64$ & Spatial distribution of logic cells & Approximates static and dynamic current demand \\
\midrule
Input & input\_switching.npy & $64 \times 64$ & Signal activity and toggle rate & Captures dynamic current fluctuations \\
\midrule
Output & labels\_ir\_drop.npy & $64 \times 64$ & Voltage drop distribution & Supervised target for pixel-wise regression \\
\bottomrule
\end{tabular}
\end{table}

\section{Synthetic Ground Truth Generation}
\label{sec:synthetic}

Accurate IR-drop labels are typically obtained using detailed physics-based solvers, which require complete design information and incur high computational cost. Such data is often unavailable during early stages of physical design. To address this limitation, this work adopts a physics-inspired synthetic labeling strategy that approximates IR-drop behavior while remaining computationally efficient.

The synthetic ground truth is generated based on the fundamental relationship between current demand and power grid resistance. Specifically, IR-drop is proportional to the product of current and resistance, where current demand is influenced by both cell density and switching activity, and resistance is inversely related to power grid strength.

The synthetic IR-drop map is computed using the following formulation:
\begin{equation}
\text{IR-drop} = \frac{\text{Cell Density} \times \text{Switching Activity}}{\text{Power Grid Strength} + \epsilon}
\label{eq:synthetic}
\end{equation}
where $\epsilon$ is a small constant added for numerical stability.

To better resemble realistic voltage drop distributions observed in practice, the resulting map is further smoothed using a spatial filtering operation, which captures both global voltage droop and localized hotspot behavior. Finally, the values are normalized to the range $[0,1]$ to ensure numerical stability during neural network training.

Although this synthetic approach does not solve the full physical equations governing power delivery networks, it preserves the key physical dependencies between resistance, current demand, and voltage drop. As a result, the generated labels provide meaningful supervision for training the deep learning model and enable effective learning of spatial IR-drop patterns. The synthetic formulation is designed for early-stage approximation and does not aim to replace signoff-grade physics solvers.

\section{Preprocessing Pipeline}
\label{sec:preprocessing}

Prior to training, all input feature maps undergo a consistent preprocessing pipeline to ensure numerical stability and effective learning. Each input map is normalized to the range $[0,1]$, which prevents scale imbalance across features and improves convergence during optimization.

For each sample, the three input maps—power grid strength, cell density, and switching activity—are stacked along the channel dimension to form a tensor of shape $(3, H, W)$. This multi-channel representation preserves spatial alignment across features and enables convolutional layers to jointly learn local and global spatial correlations.

The dataset is divided into training and validation subsets to monitor generalization performance. The training set is used to optimize network parameters, while the validation set is used to evaluate model performance on unseen samples during training. Validation loss is monitored to detect overfitting and guide model selection.

This preprocessing pipeline ensures a structured and reproducible workflow, allowing the deep learning model to focus on learning meaningful spatial patterns rather than being influenced by data scale or distribution artifacts.
\begin{center}
\begin{tabular}{c}
\textbf{Input Feature Maps} \\
(Power Grid, Cell Density, Switching Activity) \\
$\downarrow$ \\
\textbf{Preprocessing} \\
(Normalization and Tensor Stacking) \\
$\downarrow$ \\
\textbf{U-Net Convolutional Neural Network} \\
(Encoder $\rightarrow$ Bottleneck $\rightarrow$ Decoder) \\
$\downarrow$ \\
\textbf{Predicted IR-Drop Heatmap} \\
$\downarrow$ \\
\textbf{Visualization and Analysis} \\
(Hotspot Detection and Risk Assessment)
\end{tabular}
\end{center}

\section{Methodology}
\label{sec:methodology}

The proposed approach formulates IR-drop estimation as a pixel-wise regression problem, where the objective is to predict a continuous voltage drop value for each spatial location in the layout. Given multi-channel spatial input maps, the model learns a direct mapping to the corresponding IR-drop heatmap.

A CNN is used as a surrogate model to approximate the complex relationship between layout-level features and voltage drop behavior. Rather than explicitly solving physical equations during inference, the network learns this relationship from data through supervised learning. This approach is inspired by recent applications of machine learning to VLSI physical design problems \cite{wang2020mlvlsi, lee2021fastpower}.

The model is trained in an end-to-end manner using backpropagation and gradient-based optimization. All network parameters are optimized jointly to minimize the discrepancy between predicted and ground-truth IR-drop maps. This allows the model to automatically learn hierarchical spatial features without relying on handcrafted rules or explicit physical solvers.

Once trained, IR-drop prediction is performed using a single forward pass of the network, enabling fast inference suitable for early-stage design analysis. This data-driven formulation significantly reduces computation time while maintaining spatial accuracy in the predicted voltage drop distributions.

\section{Model Architecture (U-Net)}
\label{sec:architecture}

To model the spatial nature of IR-drop, a U-Net-based CNN is employed \cite{ronneberger2015unet}. U-Net is well suited for pixel-wise regression tasks where both global context and fine-grained spatial details must be preserved. Since IR-drop exhibits smooth voltage gradients as well as localized hotspots, capturing multi-scale spatial information is essential.

The architecture follows an encoder-decoder structure. The encoder progressively downsamples the input feature maps using convolutional layers, enabling the network to learn high-level representations that capture global voltage behavior. The bottleneck layer aggregates contextual information across the entire spatial region.

The decoder upsamples the encoded features back to the original resolution, reconstructing a dense IR-drop heatmap. Skip connections between corresponding encoder and decoder layers are used to transfer fine-grained spatial information directly, preventing loss of local layout details during downsampling. This design improves spatial accuracy and stabilizes training.

The network takes a multi-channel input tensor of shape $(3, H, W)$ and outputs a single-channel IR-drop map of shape $(1, H, W)$. All layers are fully convolutional, allowing the model to generalize across spatial regions without requiring fixed-size handcrafted features.

By combining hierarchical feature extraction with spatial reconstruction, the U-Net architecture enables accurate, end-to-end learning of IR-drop patterns from layout-level inputs.

\begin{figure}[t]
\centering
\includegraphics[width=0.95\textwidth]{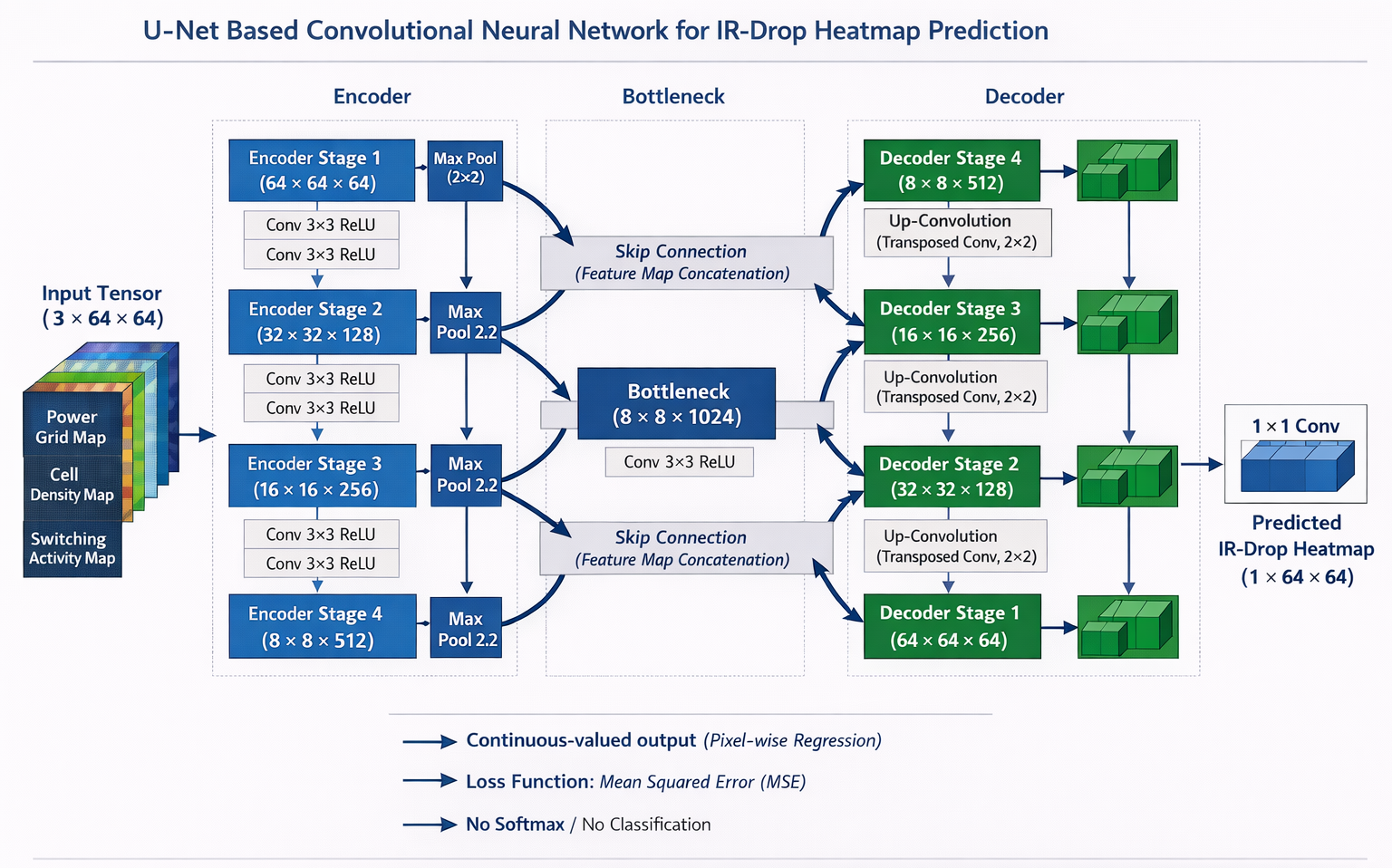}
\caption{U-Net-based CNN for IR-drop heatmap prediction. The model performs pixel-wise regression from three physical layout feature maps to a continuous IR-drop distribution using an encoder-decoder architecture with skip connections.}
\label{fig:unet}
\end{figure}

\section{Training Strategy}
\label{sec:training}

The proposed U-Net model is trained in a supervised, end-to-end manner to learn a direct mapping from physical layout feature maps to continuous IR-drop heatmaps. Validation loss monitoring and early stopping are used to reduce overfitting and improve generalization to unseen layouts.

\subsection{Loss Function}

We formulate IR-drop prediction as a pixel-wise regression problem. The model is trained using Mean Squared Error (MSE) loss:
\begin{equation}
\mathcal{L}_{\text{MSE}} = \frac{1}{HW} \sum_{i=1}^{H} \sum_{j=1}^{W} (Y_{ij} - \hat{Y}_{ij})^2
\label{eq:mse}
\end{equation}
where $Y_{ij}$ is the ground truth IR-drop value at spatial location $(i,j)$ and $\hat{Y}_{ij}$ is the predicted value.

MSE is chosen for the following reasons:
\begin{itemize}
    \item MSE penalizes large voltage prediction errors more strongly. It is well-suited for continuous-valued spatial regression. It encourages accurate hotspot magnitude estimation
\end{itemize}

This choice aligns with the objective of minimizing voltage estimation error across the entire layout.

\subsection{Optimizer}

Training is performed using the Adam optimizer \cite{kingma2014adam}, due to its:
\begin{itemize}
    \item Adaptive learning rate per parameter. Fast convergence in deep convolutional networks. Robustness to noisy gradients
\end{itemize}

Adam enables stable training without extensive manual tuning.

\subsection{Hyperparameters}

The following hyperparameters are used during training:
\begin{itemize}
    \item \textbf{Learning rate:} $1 \times 10^{-3}$
    \item \textbf{Batch size:} 8--16 (depending on GPU memory)
    \item \textbf{Number of epochs:} 50--100
    \item \textbf{Weight initialization:} He initialization
    \item \textbf{Activation function:} ReLU
    \item \textbf{Output activation:} None (linear output for regression)
\end{itemize}

Early stopping and validation loss monitoring are used to prevent overfitting.

\subsection{Training Setup}

The training configuration includes:
\begin{itemize}
    \item Input tensors are normalized to $[0,1]$. Ground-truth IR-drop maps are normalized consistently. The model is trained on synthetic but physics-inspired data. Validation loss is monitored to ensure generalization
\end{itemize}

Once trained, inference requires only a single forward pass, enabling millisecond-level prediction time.

\section{Evaluation Metrics}
\label{sec:metrics}

To evaluate the performance of the proposed model, we use metrics commonly applied in image-based regression tasks.

\subsection{Mean Squared Error (MSE)}

Mean Squared Error measures the average squared difference between predicted and ground-truth IR-drop values at each pixel:

MSE is used both as a training loss and an evaluation metric.

\subsection{Peak Signal-to-Noise Ratio (PSNR)}

PSNR evaluates the spatial similarity between predicted and reference heatmaps:
\begin{equation}
\text{PSNR} = 10 \log_{10} \left( \frac{\text{MAX}^2}{\text{MSE}} \right)
\label{eq:psnr}
\end{equation}
where MAX is the maximum possible pixel value.

PSNR provides the following insights:
\begin{itemize}
    \item Higher PSNR indicates better reconstruction quality
    \item Sensitive to structural differences in heatmaps
    \item Useful for assessing visual and spatial fidelity of IR-drop patterns
\end{itemize}

PSNR complements MSE by capturing perceptual and spatial accuracy, not just numerical error.

\subsection{Qualitative Evaluation}

In addition to numerical metrics, predicted IR-drop maps are visually compared against ground-truth heatmaps to assess:
\begin{itemize}
    \item Hotspot localization accuracy. Spatial continuity of voltage drop regions. Preservation of layout-dependent patterns
\end{itemize}

\section{Results and Analysis}
\label{sec:results}

\subsection{Quantitative Results}

The proposed U-Net-based model achieves strong regression performance on the validation dataset. Table~\ref{tab:results} summarizes the quantitative evaluation metrics.

\begin{table}[h]
\centering
\caption{Performance metrics of the proposed IR-drop estimation model}
\label{tab:results}
\begin{tabular}{lc}
\toprule
\textbf{Metric} & \textbf{Value} \\
\midrule
Mean Squared Error (MSE) & $\approx 4.9 \times 10^{-4}$ \\
Peak Signal-to-Noise Ratio (PSNR) & $\approx 33.3$ dB \\
Inference Time (per sample) & $< 10$ ms (GPU) \\
\bottomrule
\end{tabular}
\end{table}

The model is both fast and accurate. A low error score (MSE) shows it predicts voltage levels correctly at the pixel level. A high similarity score (PSNR) proves the predicted maps look almost identical to the real reference maps. Because it generates results in under 10 milliseconds, this makes the approach well suited for early-stage design exploration.

\subsection{Qualitative Analysis}

Visual inspection shows the model accurately identifies hotspots, which are dangerous areas where voltage drops too low. It correctly shows that these problems occur most in crowded areas or where components work very hard. It also captures smooth voltage changes across the chip, matching how electricity actually flows.

\subsection{Inference Efficiency}

The main advantage is speed. Traditional numerical tools are slow because they solve complex equations iteratively. This model finishes the job in milliseconds with just one forward pass. This massive speedup allows designers to explore more options and assess risks quickly.

\subsection{Generalization Behavior}

The model does not just memorize old designs. It performs well even on new layouts it has never seen before. The use of skip connections in the U-Net architecture is key to this, as they help preserve fine-grained spatial details and improve accuracy.

\section{Application and Deployment}
\label{sec:deployment}

To enable practical usage and rapid evaluation, the trained model is integrated into an interactive web-based application built using Streamlit. The live application is available at: \url{https://ir-drop-predictor.streamlit.app/}

\subsection{Interactive User Interface}

The application provides a user-friendly interface that allows designers to predict IR-drop without needing expensive simulation software or pre-existing training labels. It features two distinct modes: a Dataset Evaluation mode for testing the model on built-in samples, and an Inference-Only mode where users can upload their own custom layout maps. This is specifically designed for early-stage design when real physics-based data is not yet available.

\subsection{The Inference Workflow}

The application follows a clear, four-step pipeline to turn design data into a visual report:

\begin{enumerate}
    \item \textbf{Input Acquisition:} Three 2D maps are provided representing the power grid strength, cell density, and switching activity
    \item \textbf{Preprocessing:} The system normalizes these maps and stacks them into a single tensor for processing
    \item \textbf{Model Inference:} The trained U-Net model performs a single forward pass to predict the voltage drop
    \item \textbf{Analysis and Visualization:} The application displays a colorful heatmap of the results and calculates vital statistics, such as the total number of hotspots and the overall risk level
\end{enumerate}

\subsection{Deployment and Practical Impact}

The application is designed to be fast, flexible, and platform-independent, meaning it can run on a local computer or a cloud server. Because it only performs inference (and doesn't require constant retraining), it provides immediate feedback to engineers.

This bridge between academic research and practical work is significant because it allows for rapid estimation and the early detection of power failures. By catching these issues early, companies can significantly reduce their reliance on slow and expensive simulation cycles.

\begin{figure}[t]
\centering
\includegraphics[width=0.95\textwidth]{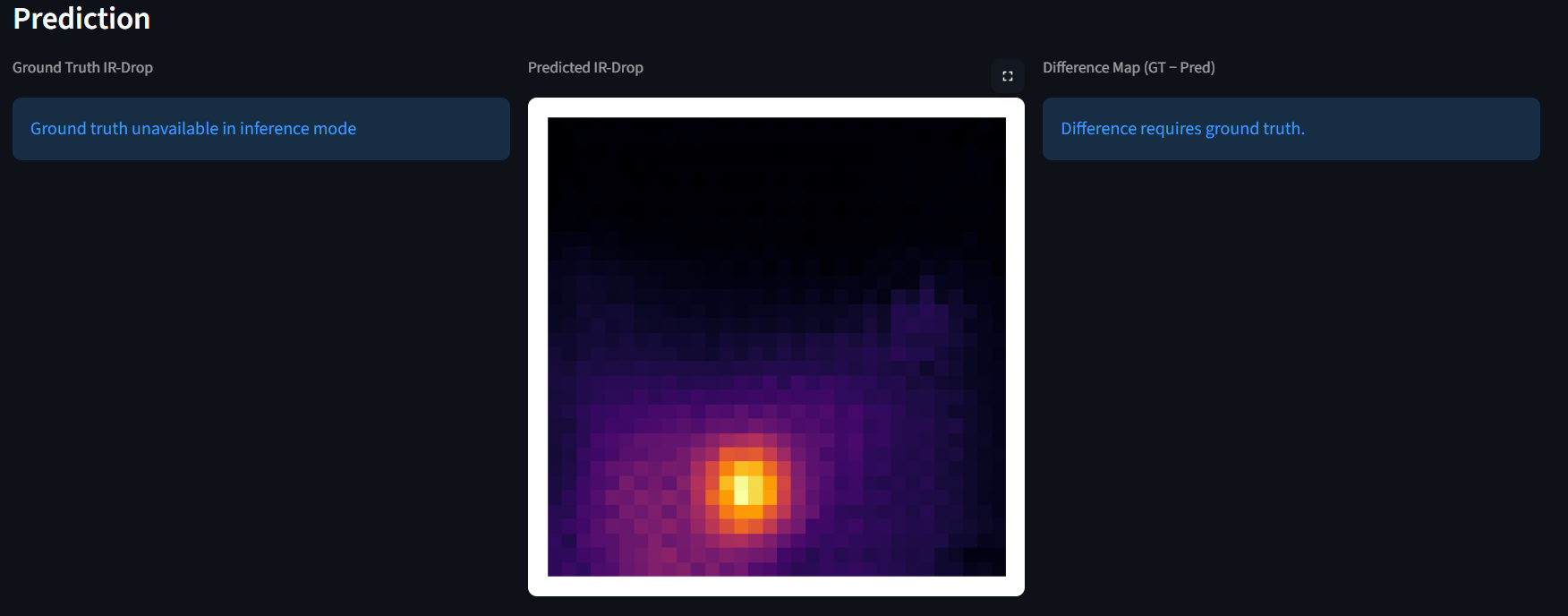}
\caption{Predicted IR-drop heatmap with spatial distribution of voltage drops and identified hotspot regions. The visualization shows maximum IR-drop of 1.0904, average IR-drop of 0.1281, with 17 hotspots identified, resulting in a HIGH RISK classification.}
\label{fig:results}
\end{figure}

\section{Challenges and Limitations}
\label{sec:limitations}

\subsection{Synthetic Data Dependency}

A key limitation of this work is the reliance on synthetically generated ground-truth IR-drop data. While the synthetic labels preserve key physical dependencies, they do not capture all real-world effects such as temperature variation, package parasitics, or transient noise. As a result, the model learns an approximation of IR-drop behavior rather than an exact physical solution.

\subsection{Generalization to Real Designs}

Although the model demonstrates strong performance on held-out validation data, generalization to unseen, real industrial layouts remains a challenge. Differences in technology nodes, grid topology, and workload characteristics may impact prediction accuracy. Additional training data from diverse design scenarios would further improve robustness and reduce potential domain shift.

\subsection{Threshold Calibration for Risk Classification}

The definition of IR-drop risk levels (e.g., hotspot thresholds) is currently based on fixed empirical values. In practice, acceptable IR-drop limits vary across designs and operating conditions. Future implementations may require adaptive or design-specific threshold calibration to align predictions with specific power integrity constraints.

\subsection{Model Interpretability}

Like most deep learning models, the proposed U-Net operates as a black-box predictor. While it provides accurate spatial predictions, it does not explicitly explain which input features contribute most to specific IR-drop patterns. This limits direct interpretability compared to physics-based solvers.

\subsection{Scope of Applicability}

The proposed approach is intended for early-stage estimation and rapid exploration, not for final signoff. It should be viewed as a complementary tool that provides fast feedback prior to running detailed simulation-based analysis.

\section{Conclusion}
\label{sec:conclusion}

This work presents a deep learning-based approach for fast IR-drop estimation using a U-Net CNN. By reformulating IR-drop analysis as a pixel-wise regression problem, the proposed model learns a direct mapping from physical layout feature maps to continuous IR-drop heatmaps.

The results demonstrate that the model can accurately capture spatial voltage drop patterns and identify potential hotspot regions while achieving orders-of-magnitude faster inference compared to traditional simulation-based approaches. This makes the method well-suited for early-stage power integrity analysis, where rapid feedback is essential for iterative design exploration.

Rather than replacing physics-based solvers, the proposed approach serves as a data-driven surrogate model that complements existing analysis flows. It enables designers to quickly assess power integrity risk and prioritize design refinements before running detailed and computationally expensive simulations.

Overall, this work highlights the potential of deep learning techniques to accelerate complex physical analysis tasks and supports the growing role of machine learning in design automation workflows.

The implementation of the proposed model, dataset generation scripts, and the interactive inference application are publicly available at: \url{https://github.com/riteshbhadana/IR-Drop-Predictor}. Future work includes validation using industrial signoff datasets and integration with physical design workflows.

\bibliographystyle{IEEEtran}
\bibliography{references}

@inproceedings{ronneberger2015unet,
  title={U-Net: Convolutional Networks for Biomedical Image Segmentation},
  author={Ronneberger, Olaf and Fischer, Philipp and Brox, Thomas},
  booktitle={Medical Image Computing and Computer-Assisted Intervention--MICCAI 2015},
  pages={234--241},
  year={2015},
  organization={Springer}
}

@article{kingma2014adam,
  title={Adam: A Method for Stochastic Optimization},
  author={Kingma, Diederik P and Ba, Jimmy},
  journal={arXiv preprint arXiv:1412.6980},
  year={2014}
}

@article{wang2020mlvlsi,
  title={Machine Learning for VLSI Physical Design: A Survey},
  author={Wang, Siting and Li, Yibo and Fang, Hui and Pan, David Z},
  journal={IEEE Transactions on Computer-Aided Design of Integrated Circuits and Systems},
  volume={40},
  number={8},
  pages={1650--1670},
  year={2020},
  publisher={IEEE}
}

@inproceedings{lee2021fastpower,
  title={Fast Power Integrity Analysis Using Machine Learning Techniques},
  author={Lee, Kyungwook and Jeong, Seokho and Kim, Jihwan},
  booktitle={Proceedings of the Design Automation Conference (DAC)},
  pages={1--6},
  year={2021}
}

\end{document}